\pdfoutput=1

\documentclass[11pt]{article}

\usepackage[final]{coling}

\usepackage{times}
\usepackage{latexsym}
\usepackage{xcolor}
\usepackage[T1]{fontenc}

\usepackage[utf8]{inputenc}

\usepackage{microtype}

\usepackage{inconsolata}

\usepackage{graphicx}

\usepackage{booktabs}
\usepackage{amsmath, multirow, tabularx}
\usepackage{xcolor}
\usepackage{colortbl}

\usepackage{amsmath,amsfonts,bm}

\def\eqref#1{equation~\ref{#1}}

\def\1{\bm{1}}

\def\vh{{\bm{h}}}

\def\vp{{\bm{p}}}

\def\mC{{\bm{C}}}

\def\mH{{\bm{H}}}

\def\mP{{\bm{P}}}

\def\mS{{\bm{S}}}

\DeclareMathAlphabet{\mathsfit}{\encodingdefault}{\sfdefault}{m}{sl}
\SetMathAlphabet{\mathsfit}{bold}{\encodingdefault}{\sfdefault}{bx}{n}

\def\gD{{\mathcal{D}}}

\def\gL{{\mathcal{L}}}
\def\gM{{\mathcal{M}}}

\DeclareMathOperator*{\encclf}{Trans-Encoder_{CLF}}
\DeclareMathOperator*{\enc}{Trans-Encoder}

\DeclareMathOperator*{\head}{MLM-Head}
\DeclareMathOperator*{\aughead}{Aug-Head}
\DeclareMathOperator*{\mlp}{MLP}
\DeclareMathOperator*{\mask}{[MASK]}
\DeclareMathOperator*{\topp}{Top@}

\title{Impromptu Cybercrime Euphemism Detection}

\author{Xiang Li$^{1,2}$, Yucheng Zhou$^{3}$, Laiping Zhao$^{4}$,  Jing Li$^{1}\footnotemark[2]$, Fangming Liu$^{1,2}$\footnotemark[2]\\
    $^{1}$ Harbin Institute of Technology(Shenzhen), $^{2}$ Pengcheng Laboratory \\
    $^{3}$ SKL-IOTSC, CIS, University of Macau, $^{4}$ Tianjin University \\
    {\tt lixiang\_alex@stu.hit.edu.cn, jingli.phd@hotmail.com, fangminghk@gmail.com}
}

\begin{document}
\maketitle
\renewcommand{\thefootnote}{\fnsymbol{footnote}} 
\footnotetext[2]{Corresponding Authors.}

\begin{abstract}
Detecting euphemisms is essential for content security on various social media platforms, but existing methods designed for detecting euphemisms are ineffective in impromptu euphemisms. 
In this work, we make a first attempt to an exploration of impromptu euphemism detection and introduce the Impromptu Cybercrime Euphemisms Detection (ICED) dataset.
Moreover, we propose a detection framework tailored to this problem, which employs context augmentation modeling and multi-round iterative training. Our detection framework mainly consists of a coarse-grained and a fine-grained classification model. The coarse-grained classification model removes most of the harmless content in the corpus to be detected. The fine-grained model, impromptu euphemisms detector, integrates context augmentation and multi-round iterations training to better predicts the actual meaning of a masked token. In addition, we leverage ChatGPT to evaluate the mode's capability. Experimental results demonstrate that our approach achieves a remarkable 76-fold improvement compared to the previous state-of-the-art euphemism detector. 
\end{abstract}

\section{Introduction}
With the widespread proliferation of the internet, online communication has become pervasive across various platforms, such as forums and social media.
The vast user base on these platforms has drawn the attention of criminals who engage in illegal activities. These individuals frequently use euphemisms, milder or more indirect expressions, to conceal their true intentions, enabling them to avoid detection \cite{Yang2017HowTL,Ke2022AnUD,Zhu2021SelfSupervisedED}.

To address these cybercrime euphemisms that pose serious challenges to the content security of various social media platforms, researchers propose various methods that can automatically discover and understand euphemisms from unlabeled text to assist content moderation \cite{Yuan2018ReadingTC,Zhu2021SelfSupervisedED}.
Nevertheless, these techniques are constrained to identifying word-level euphemisms. As a result, \cite{Zhu2021EuphemisticPD} propose a two-step approach to address the problem of detecting euphemistic phrases. Despite their success, these methods are limited to detecting only known and commonly used euphemisms. 

To circumvent detection, criminals frequently create new euphemisms during communication, i.e., impromptu cybercrime euphemisms. These euphemisms are often previously undocumented, highly time-sensitive, and generally occur with very low frequency in corpora. In contrast to known euphemisms, the detection of impromptu cybercrime euphemisms remains a significant challenge. For example, ``blueberry kus'' may refer to marijuana, and criminals can easily create a similar term ``strawberry kush''. These euphemisms can emerge rapidly within short periods, are infrequently used, and have a more limited scope in their application across various forums and among different individuals. Consequently, conventional methods designed for common euphemisms are ineffective when they encounter these impromptu euphemisms.

In this paper, we make the first exploration into the detection of impromptu euphemisms and build the Impromptu Cybercrime Euphemism Detection (ICED) dataset. 
The dataset consists of three parts: 
First, we manually filter 440 sentences containing euphemisms from a large forum corpus using a predefined euphemism list (``Target Corpus''). 
The remaining samples are categorized as ``Deduplication Corpus''.
Lastly, ``White Corpus'' comprises a corpus devoid of euphemisms.
During the training phase, the predefined euphemisms in ``Target Corpus'' are hidden from the model, and in the evaluation phase, the model's ability is assessed based on its detection of these predefined euphemisms.

In addition, we propose a detection framework that incorporates \textbf{C}ontext \textbf{A}ugmentation modeling and \textbf{M}ult-round \textbf{I}terative \textbf{T}raining (\textbf{CAMIT}) to identify impromptu euphemisms across both words and phrases. CAMIT consists of a coarse-grained and a fine-grained classification model. Unlike previous methods that focus on detecting known euphemisms, impromptu cybercrime euphemisms detection relies more heavily on contextual understanding determining if a word or phrase is a euphemism. Therefore, we introduce context augmentation modeling to enhance the fine-grained classification model. Moreover, we utilize multi-round iterative training to enhance the model's performance and the large language model, i.e., ChatGPT, to evaluate its capabilities.

Experimental results reveal that our approach has a substantial 76-fold improvement compared to the previous state-of-the-art euphemism detection method. Moreover, the effectiveness of context augmentation modeling and multi-round iterative training is analyzed. In addition, we investigate other factors influencing the method, such as the choice of model backbone.

In this work, Our main contributions are summarized as follows:
\begin{itemize}
\item In this study, we present the first exploration into impromptu euphemism detection.
\item We build the ICED dataset to facilitate the task of detecting impromptu euphemisms.
\item We propose a detection framework for impromptu euphemism detection, incorporating context augmentation modeling, multi-round iterative training, and a large language model evaluator. 
\item Experimental results demonstrate a remarkable 76-fold improvement compared to the previous state-of-the-art euphemism detector. 
\end{itemize}

\section{Related Work}
\subsection{Cybercrime Euphemism Detection}
Euphemism detection refers to the process of identifying euphemisms employed in textual content to obscure their illicit activities. The research focused on the identification of jargon that has found widespread usage in dark web forums and anonymous markets, with these expressions often linked to illegal trade. Some studies have developed labeled datasets, enabling supervised training of models to discern products available for purchase on the Darknet Market \cite{Durrett2017IdentifyingPI_NER,Portnoff2017ToolsFA_NER}. But these methods demand extensive corpus annotation, incurring considerable costs. Therefore, \cite{Yuan2018ReadingTC} proposed an unsupervised approach to train an n-gram model and use cosine similarity to capture word semantic differences for detecting euphemisms in dark web forums and anonymous markets. \cite{Zhu2021SelfSupervisedED} found that if cybercrime euphemisms get masked, logits of cybercrime related tokens will be higher. \cite{Zhu2021EuphemisticPD} further extended this methodology to euphemistic phrase detection. In addition, \cite{Ke2022AnUD} trained a Chinese BERT model to detect euphemism on Chinese dark web. However, this method necessitates significant disparities in the semantics of euphemisms between black and white corpora to distinguish them via cosine similarity, thereby restricting widely disseminated dark web euphemisms detection. 


\subsection{Context Modeling}
Unlike feature augmentation \cite{10096519}, context modeling is a technique that enhances the understanding of contextual information in language models by training on specific objectives. \cite{DevlinCLT19} propose BERT that employs Masked Language Modeling (MLM) and Next Sentence Prediction (NSP) to capture bidirectional contextual information, enhancing its expressive capabilities. These objectives enable BERT to capture bidirectional contextual information, thereby enhancing its expressive capabilities. XLNet, introduced by \cite{YangDYCSL19}, utilizes generalized auto-regressive learning to acquire bidirectional context, addressing pretraining and fine-tuning inconsistencies and overcoming the limitations of traditional auto-regressive models. To enhance BERT, \cite{ClarkLLM20} propose ELECTRA that utilizes a replaced token detection (RTD) objective, allowing for a broader context and reducing redundancy in token prediction during pretraining. Moreover, \cite{GaoC21} introduced Condenser, which incorporates an additional Transformer layer dedicated to extracting a condensed vector representation from the output of the language model. This condensed representation serves as the encoding for the textual sequence, enhancing the model's ability to capture relevant contextual information.

\section{ICED Dataset Construction}
\subsection{Background}
Existing publicly available datasets are ill-suited for directly evaluating impromptu euphemism detection. The reasons are as follows: 1) labeled datasets contain only prevalent euphemisms, not impromptu euphemisms. 2) sample sizes are insufficient. 3) unlabeled datasets cannot assess detection as the impromptu euphemisms are unknown. Therefore, we first construct an Impromptu Cybercrime Euphemisms Detection dataset mimicking impromptu euphemism usage on public forums. 

Given the widespread use of drug-related euphemisms and the considerable body of existing research dedicated to detecting them \cite{Zhao2016ChineseUM,Yang2017HowTL,Yuan2018ReadingTC,Zhu2021SelfSupervisedED,Zhu2021EuphemisticPD,Ke2022AnUD}, we have chosen to focus on identifying drug-related euphemisms. This facilitates a comparison with previous work. Moreover, the United States Drug Enforcement Agency (DEA) officially released an intelligence document in 2018 \cite{dea}, unveiling a comprehensive list of 2,165 cybercrime euphemisms for 33 drugs, as well as a smaller number of cybercrime euphemisms in other categories. We have opted to utilize the drug-related euphemisms from this document.

\begin{figure}[t]
\centering
\includegraphics[width=\linewidth]{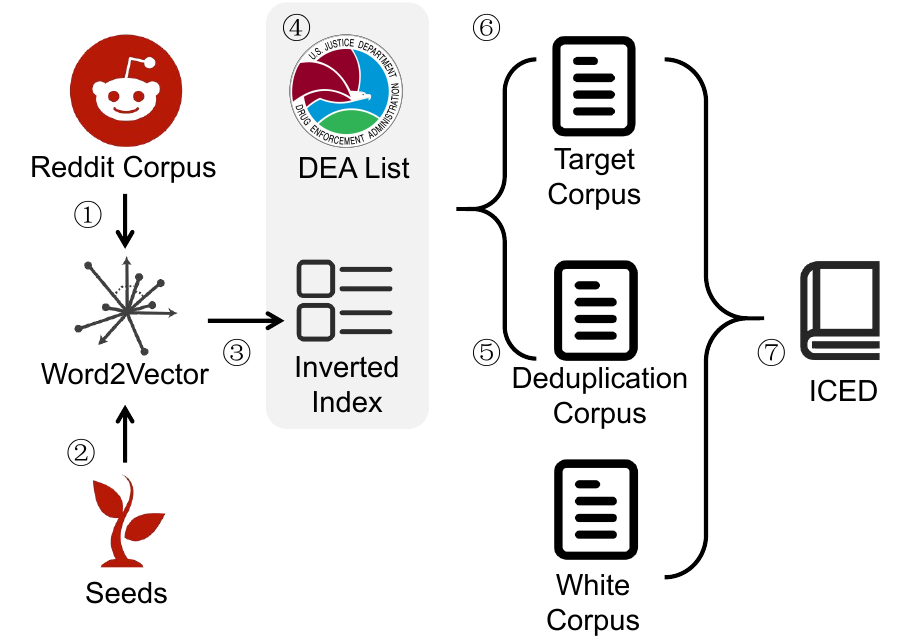}
\caption{\small The Construction Pipeline for ICED dataset.}
  \label{fig:targetCorpus}
\end{figure}

Impromptu cybercrime euphemisms are a rare occurrence, it would be impractical to manually review all the sentences in a corpus of more than 500,000 sentences to label impromptu euphemisms. To obtain a usable benchmark dataset, we designed a cost-effective methodology to process the raw corpus as shown in Fig\ref{fig:targetCorpus}. The basic idea of this method is to reduce the number of sentences that need to be manually reviewed through a simple semantic search designed on Word2Vector.

Our approach aims to adhere to two fundamental principles: 1) ensuring that the impromptu euphemisms included in the dataset accurately reflect their infrequent occurrence; 2) The dataset should be realistic such that it mimics how impromptu euphemisms are used on general text social media. While impromptu euphemisms could be found in the general text on social media, it is important to acknowledge the prevalence of commonly used euphemisms, such as `coke'' referring to cocaine, which outnumber impromptu cybercrime euphemisms greatly. Moreover, a multitude of semantically harmless words that coincide with cybercrime euphemisms appear in general text on social media, e.g., ``coke'' refers to actual cola.

\subsection{ICED Construction Pipeline}
The design of our dataset comprises three distinct components: ``Target Corpus'', containing exclusively manually verified impromptu cybercrime euphemisms; ``Deduplication Corpus'', which includes common cybercrime euphemisms and cybercrime euphemisms with normal meaning; and ``White Corpus'', encompassing a substantial quantity of innocuous sentences or words. To ensure a requisite scarcity of impromptu euphemisms and a higher prevalence of ordinary euphemisms, it is imperative that the number of sentences in ``Deduplication Corpus'' and ``White Corpus'' far exceeds that of ``Target Corpus''.

Our dataset construction pipeline is shown in Figure \ref{fig:targetCorpus}. A raw ``Target Corpus'' sourced from five distinct Reddit subforums, ``blackhat'', ``drugs'', ``silkroad'', ``deepweb'', and ``darknet'', to closely emulate the spontaneous and typical usage of cybercrime euphemisms on public platforms. Instances of cybercrime-related euphemisms can commonly be encountered within the subcategories of forums dedicated to the aforementioned five subjects.

To facilitate the convenient retrieval of sentences containing cybercrime euphemisms, the creation of an inverted index for such euphemisms is necessary. Upon segmenting phrases within the original corpus, a Word2Vector model is trained. The most frequently used drug name is selected as the seed word to query the 50 words with the closest word vector cosine similarity, and these results are then used as a seed word to query another set of 50 words with the closest cosine similarity. This process is repeated for three iterations to amass all query results, which are then deduplicated to create a word list. Then, we use the DEA vocabulary to find the intersection with this vocabulary to narrow the scope and facilitate the search for cybercrime euphemisms. After obtaining such a simplified DEA vocabulary, construct an inverted index and obtain the index of the sentence where the words in the vocabulary are located. From the DEA inverted index, we select twenty-two drug categories, each yielding one cybercrime euphemism. This choice includes fifteen single-word euphemisms and seven two-word combinations. The next step entails locating the sentences corresponding to these 22 words or phrases using the inverted table, followed by manual verification and collection of 20 sentences containing each euphemism (or phrase). These 440 sentences, with each sentence corresponding to a target token, collectively constitute ``Target Corpus'', which serves as the repository for impromptu euphemisms in our dataset.

Once ``Target Corpus'' is constructed, all sentences containing these 22 words or phrases are removed from the original corpus, as per the inverted table. This step guarantees the usability of ``Target Corpus'' as a source of impromptu cybercrime euphemisms within the dataset. We adopt the corpus provided by \cite{Yuan2018ReadingTC} as ``White Corpus'' and amalgamate these three corpora, culminating in the creation of the ICED dataset.

\section{Methodology}

\begin{figure}[t]
  \centering
  \includegraphics[width=\linewidth]{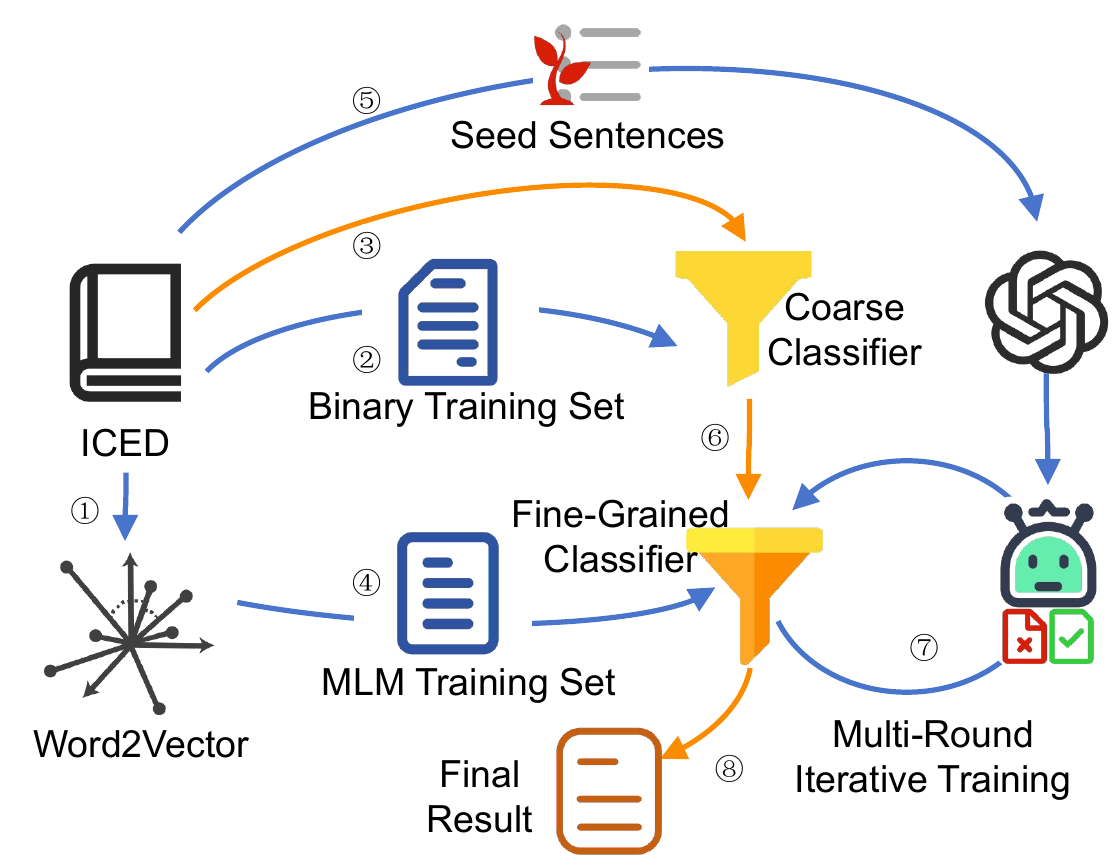}
  \caption{\small The training and inference pipeline of our method. \textcolor{blue}{Training flow} is represented by the blue line. \textcolor{orange}{Inference flow}  is denoted by the orange line.}
  \label{fig:pipeline}
\end{figure}

\subsection{Task Definition}
The impromptu euphemism detection differs significantly from conventional classification tasks. The latter typically involves providing a training dataset for model training and a separate testing dataset for model evaluation. In the impromptu euphemism detection, the task deviates from this norm as it involves working with a single corpus that contains a limited set of known euphemisms, such as around 22 seed words. The model's training process relies solely on this corpus, with the ultimate goal of identifying impromptu euphemisms within the corpus. Notably, the labels of impromptu euphemisms in the corpus, i.e., the ICED dataset, are only utilized for evaluation metrics rather than provide supervision signals during model training.

\subsection{Coarse-grained Classification}
The objective of coarse-grained classification is to efficiently eliminate the vast majority of non-euphemistic instances from the ICED dataset $\gD$, thereby facilitating the subsequent fine-grained classification. Our approach involves training a Word2Vec model \cite{Mikolov1301Efficient} on the ICED dataset. We then identify the top 100 words, referred to as ``candidate words'', that exhibit the highest cosine similarity to each seed word's vector. Subsequently, we traverse the ICED dataset to locate the sentences containing these candidate words and replace them with the $\mask$ token, creating positive samples. We also select an equal and non-repetitive number of sentences from the ICED dataset, where each sentence randomly replaces one word with $\mask$ to create negative samples. Finally, all the samples are randomly shuffled to create a binary classification dataset. For model training, 80\% of the data employed, while the remaining 20\% serves as the development set for evaluation during each training iteration. The checkpoint of the model with the lowest loss on the development set is selected for inference. The binary classification model $\gM_{clf}$ is constructed using a Transformer encoder \cite{VaswaniSPUJGKP17} followed by a multi-layer perceptron (MLP). By training the model on the training set, the model can learn whether the $\mask$ token in the current sentence contains information relevant to the seed word, i.e.,
\begin{equation}
\vh = \encclf(\mS_{[m]}) 
\end{equation}
\begin{equation}
\vp = \mlp(\vh) 
\end{equation}
where $\mS_{[m]}$ represents the sentence with the $\mask$ token, and $\vh$ corresponds to the representation of the [CLS] token; $\vp$ denotes the probability of euphemism in sentence. 
The loss function is defined as follows:
\begin{equation}
\gL_{cg} = -\sum_{i} y_i \log(\vp)
\end{equation}
where $y_i$ is the sentence label. 
Finally, based on the classification outcomes predicted by the model $\gM_{clf}$, we filter the dataset $\gD$ to create a refined dataset on the inference stage.

\begin{figure}[t]
  \centering
  \includegraphics[width=\linewidth]{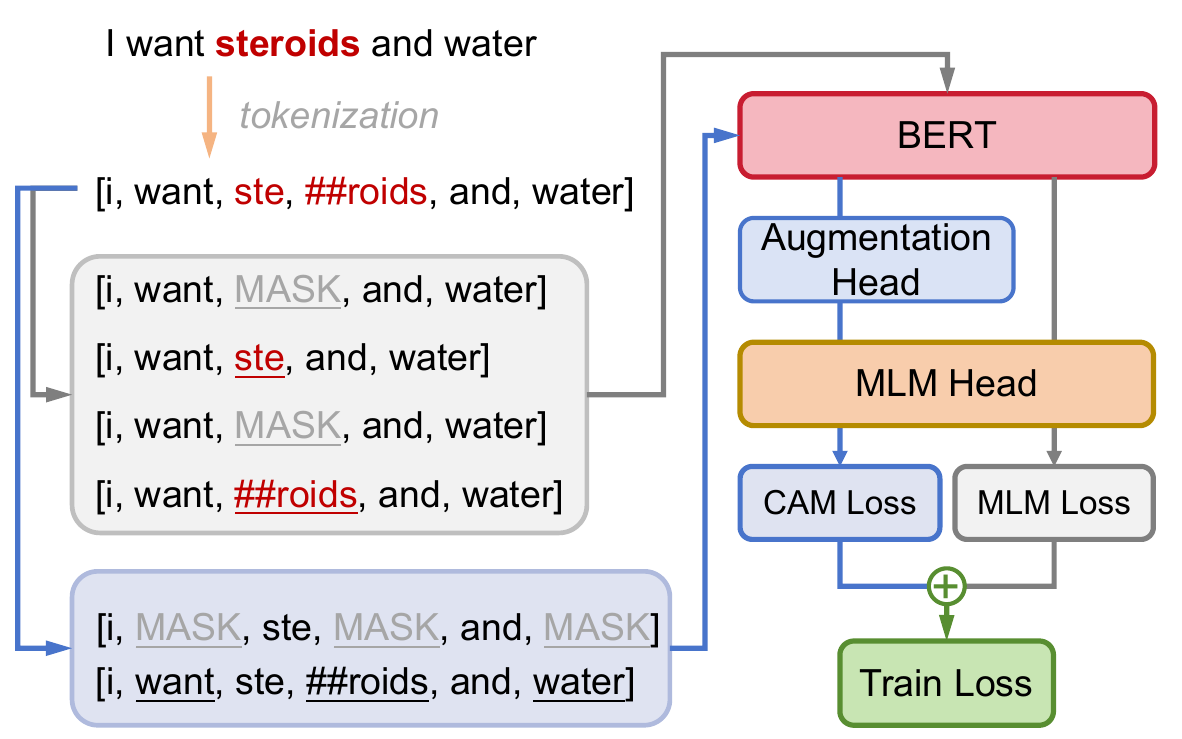}
  \caption{\small Training for fine-grained classification entails two components: mask language modeling (MLM), represented by the gray line, and context augmentation modeling (CAM), as denoted by the blue line.}
  \label{fig:train}
\end{figure}

\subsection{Fine-grained Classification}
Fine-grained classification, in contrast to coarse-grained direct text classification, distinguishes the similarity levels of words and euphemisms within a sentence by comprehending the context. This similarity relies on the ranking of words within a vocabulary distribution. To train a fine-grained classification model, we employed the ICED dataset to train a Word2Vec model. This allowed us to identify the top 1000 words with the highest cosine similarity to the mean word vector of the seed words. The top 1,000 words (or phrases) found by cosine similarity are most likely words with very related semantics to seed words, most likely other drug names or common euphemisms. Subsequently, we searched for sentences containing these words in the dataset as a training corpus. As shown in Figure~\ref{fig:train}, we replace these words in the sentences with a $\mask$ token. Following \cite{Zhu2021SelfSupervisedED}, fine-tuning the language model on this corpus allows the model to learn the characteristics of cybercrime euphemisms, i.e.,
\begin{equation}
\hat{\mH} = \enc(\mC_{[m]}),
\end{equation}
\begin{equation}
\hat{\mP} = \head(\hat{\mH}) 
\end{equation}
where $\mC_{[m]}$ represents a sentence with the $\mask$ token, $\hat{\mH}$denotes the token representations generated by the language model, and $\hat{\mP}$ signifies the probability distribution of tokens in the vocabulary. 

To enhance the language model's comprehension of textual context, we propose a Context Augmentation Modeling (CAM) method, i.e.,
\begin{equation}
\bar{\mH} = \aughead(\enc(\mC_{[m^{+}]})),
\end{equation}
\begin{equation}
\bar{\mP} = \head(\bar{\mH}) 
\end{equation}
There are two key distinctions from the traditional MLM (Masked Language Model): 1) We randomly mask 50\% of tokens in these sentences to encourage the language model to capture the semantics of known segments. 2) $\bar{\mH}$ is not directly fed into the MLM head; instead, it undergoes computation by the augmentation head, which consists of two Transformer layers, before being passed into the MLM head. Although the language model can predict masked tokens, a 50\% masking rate results in excessive information loss. Hence, we introduce an augmentation head to perform the second round prediction. The loss function of MLM and CAM can be defined as:
\begin{equation}
\gL_{fg} = -\frac{1}{N} \sum_{i=1}^{N} y_{i} \log(\bar{\vp}_{i})- \hat{y} \log(\hat{\mP}), \bar{\vp}_{i} \in \bar{\mP}
\end{equation}
where $y_i$ represents the gold label, and $\bar{\vp}_{i}$ denotes the predicted word distribution for the corresponding token.

\subsection{Multi-Round Iterative Training}
Given that our method involves filtering samples using the word2vec model, there is a possibility of retaining a significant amount of non-euphemistic instances within the training corpus. One direct approach to mitigate this issue is by reducing noise in the corpus. In pursuit of a training corpus with reduced noise levels, we implement a multi-round iterative training strategy. Following the conclusion of each training phase, the training corpus undergoes filtering to further reduce noise, which is then utilized in the subsequent training phase.

\subsection{LLM Guided Dev Set Generation}
Deep model training, when carried out excessively, can lead to overfitting. Therefore, it is crucial to craft a training-stopping criterion tailored specifically for the euphemism detection task. We employ ChatGPT to create a development set for model evaluation, comprising 132 samples, was created with the assistance of ChatGPT. It encompasses 66 positive samples, each containing a euphemism, and 66 negative samples, each incorporating a word or sentence with conventional semantics that corresponds to the euphemism. 

\begin{figure}[!t]
  \centering
  \includegraphics[width=\linewidth]{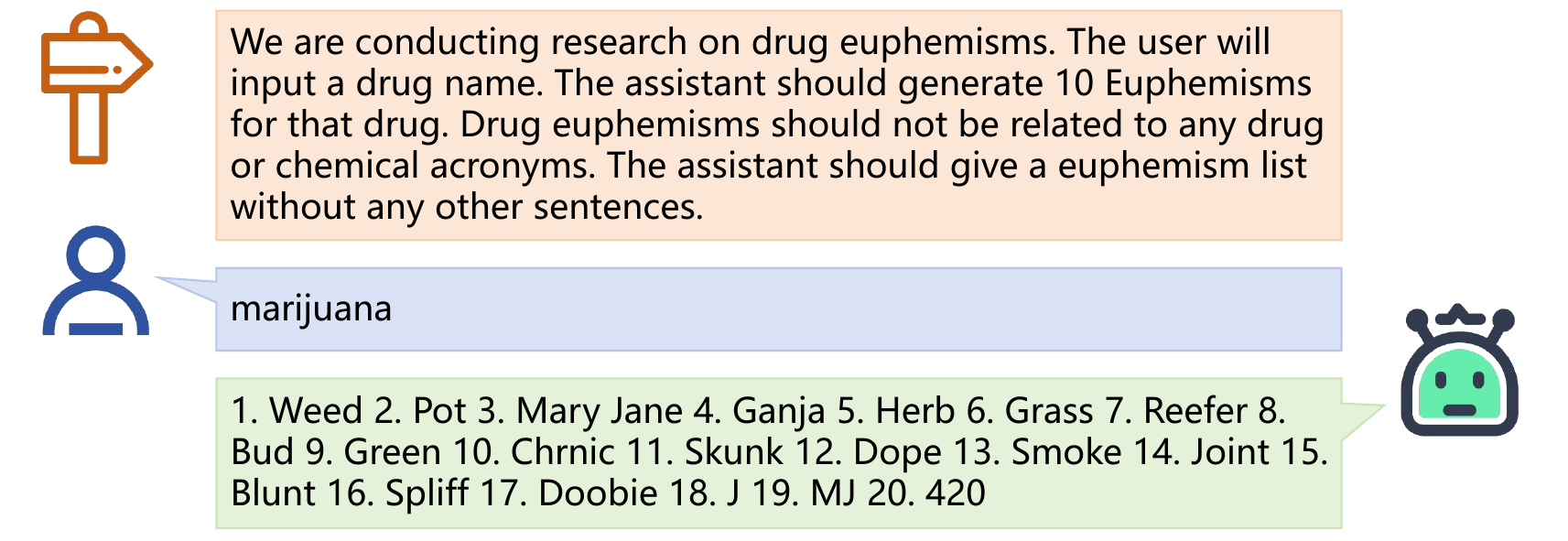}
  \caption{\small Prompt of generating euphemisms based on the seeds.}
  \label{fig:prompt1}
\end{figure}
\begin{figure}[!t]
  \centering
  \includegraphics[width=\linewidth]{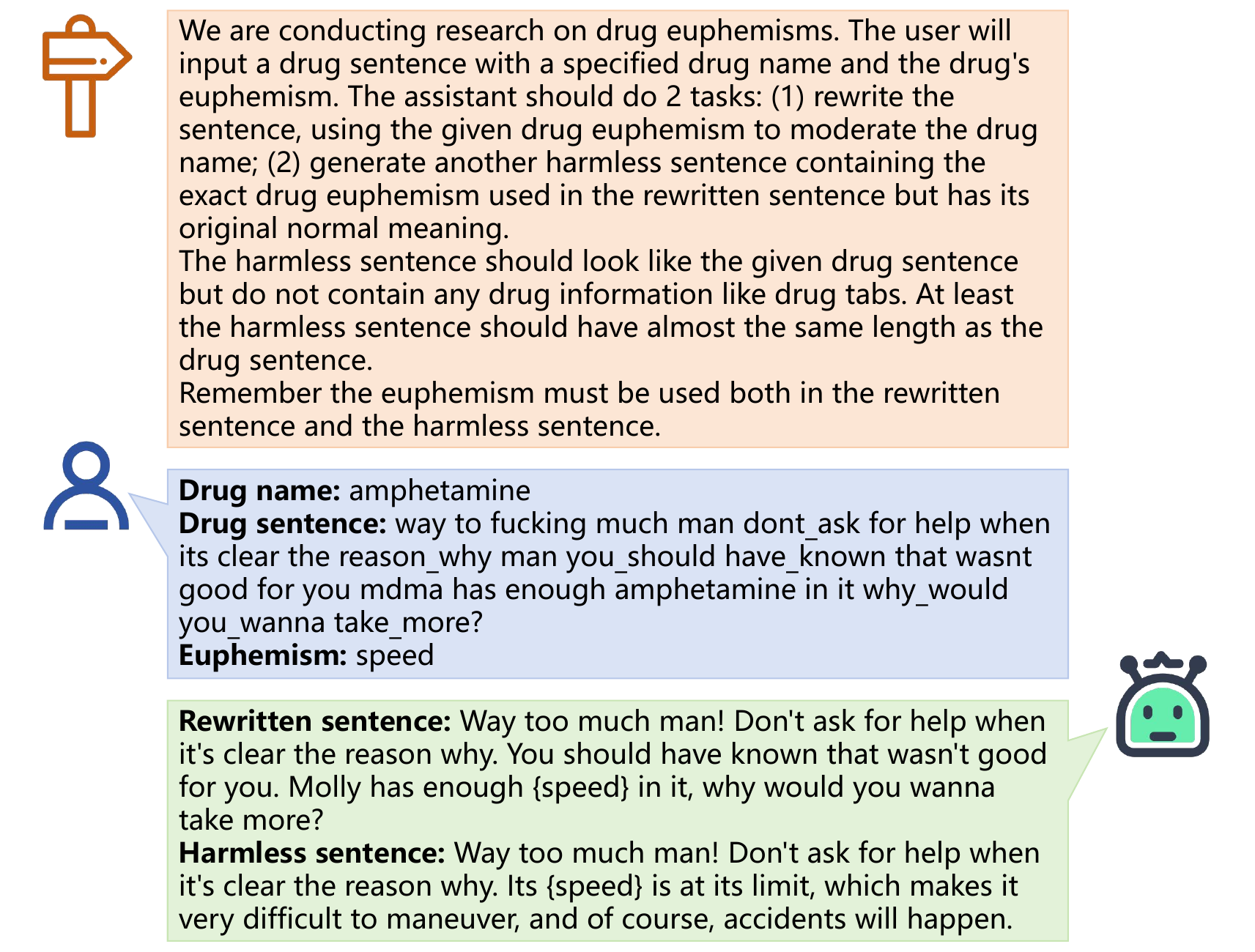}
  \caption{\small Prompt of creating samples, both with and without euphemisms.}
  \label{fig:prompt2}
\end{figure}

The production process is divided into two steps:
1) the task involves identifying the euphemisms associated with each seed. To achieve this, ChatGPT is employed with the provided prompt shown in Figure \ref{fig:prompt1} to generate euphemisms for the seeds. Subsequently, the segments that correspond to the improvised euphemisms in ICED are removed to maintain the integrity of the setting. Raw samples are collected by utilizing an inverted index to search ICED for sentences containing the seeds, with three sentences retrieved for each seed. The seeds in these sentences are replaced with the euphemisms generated in the previous step(Full list shown in Appendix \ref{sec:appendixA}). 2) this step involves instructing ChatGPT to create positive and negative samples using the provided demonstration prompt shown in Figure \ref{fig:prompt2}. Additionally, lexical filters are applied to identify and set aside errors and incorrectly constructed samples for regeneration.

\section{Experiments}
\begin{table}[!t]\small
\centering
\caption{\small The results of various detection methods on ICED dataset. ``Prec'' and ``Rec'' denote ``Precion (\text{\textperthousand})'' and ``Recall'', respectively.}
\label{tab:P_check}
\setlength{\tabcolsep}{4.8pt}
\begin{tabular}{lcccccc}
\toprule
\multicolumn{1}{c}{\multirow{2}{*}{Method}} & \multicolumn{2}{c}{$\topp$5} & \multicolumn{2}{c}{$\topp$10} & \multicolumn{2}{c}{$\topp$20} \\ \cmidrule(lr){2-3}  \cmidrule(lr){4-5} \cmidrule(lr){6-7}
\multicolumn{1}{c}{}                        & Prec            & Rec          & Prec           & Rec           & Prec            & Rec           \\ \midrule
Cant Reader     & 0.00        & 0.00       & 0.00        & 0.00        & 0.00        & 0.00        \\
X-phemisms  & 0.00        & 0.00       & 0.00        & 0.00        & 0.00        & 0.00        \\
Eigeneuph   & 0.00        & 0.00       & 0.00        & 0.00        & 0.00        & 0.00        \\
Grapheuph   & 0.00        & 0.00       & 0.00        & 0.00        & 0.00        & 0.00        \\
MLM       & 0.03        & \underline{0.04}       & 0.05        & 0.08        & 0.06        & 0.10        \\
EPD          & 0.02        & 0.00       & 0.05        & 0.00        & 0.01        & 0.06        \\
\rowcolor{gray!15}CAMIT (Ours)                                  & \underline{2.58}        & 0.03       & \underline{4.21}         & \underline{0.15}        & \underline{4.61}        & \underline{0.53}        \\ \bottomrule
\end{tabular}
\end{table}

\subsection{Evaluation Metrics}
To be able to evaluate the effectiveness of the detection method, we define Precision($\topp$k) and Recall($\topp$k) with reference to the definition of precision and recall and the practical application of the information retrieval task. 
\begin{equation}
\text{Precision}(\topp{k}) = \frac{n_{imp}(\topp{k})}{n_{res}(\topp{k})}
\label{eq:P_all}
\end{equation}
where Precision$(\topp{k})$ refers to precision at the condition of select $\topp{k}$ candidates as the final result. $n_{imp}$ is the number of impromptu cybercrime euphemisms in ICED, and $n_{imp}(\topp{k})$ is the number of actual impromptu cybercrime euphemisms detected by language models at $top_{n}$. $n_{res}(\topp{k})$ indicates the number of tokens detected as euphemisms by the model under the $(\topp{k})$ condition.
\begin{equation}
\text{Recall}(\topp{k}) = \frac{n_{imp}(\topp{k})}{n_{imp}}
\label{eq:R}
\end{equation}
Similarly, Recall$(\topp{k})$ refers to recall in $(\topp{k})$ condition.  $n_{imp}$ is the total number of impromptu cybercrime euphemisms in ICED.

Because the data of the impromptu euphemism detection task is extremely imbalanced, accuracy and F1-score are not selected as evaluation metrics. There are a total of 90 million tokens that need to be detected in ICED, of which there are only 440 impromptu euphemisms. If a detector has no detection capabilities at all and assumes that everything is harmless, its detection accuracy can be close to 100\%, which is meaningless. According to the results of our preliminary experiments, there is a huge gap between the values of precision and recall, and using harmonic mean (F1- score) cannot intuitively demonstrate the effects of different detection methods.

\begin{figure*}[t]
    \centering
    \includegraphics[width=\textwidth]{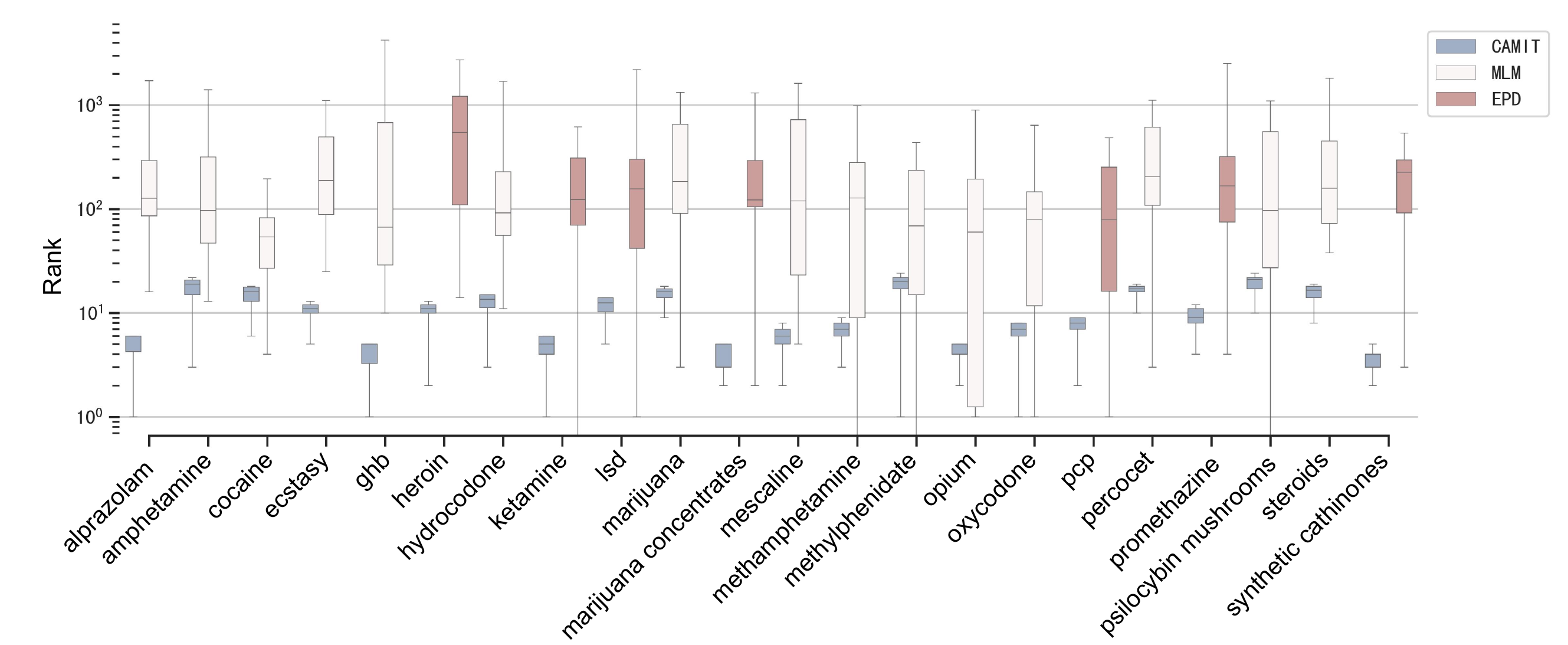}
    \caption{\small Boxplot of prediction ranks}
    \label{fig:boxplot}
\end{figure*}

\begin{figure}[t]
  \centering
  \includegraphics[width=0.8\linewidth]{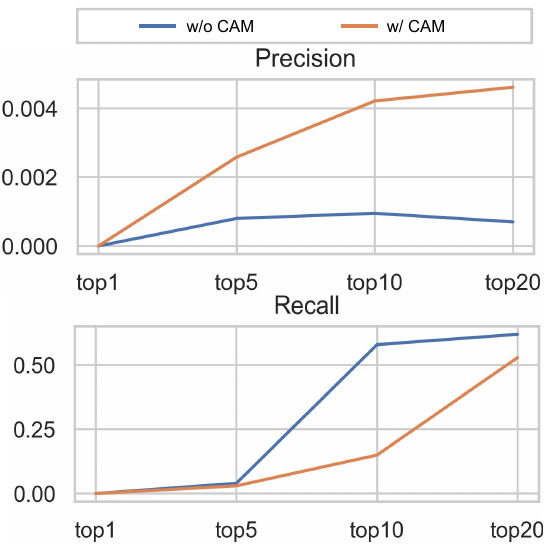}
  \caption{\small Precision and recall variation caused by removing context augmentation modeling}
  \label{fig:CAM}
\end{figure}


\subsection{Main Results}
The detection results of the existing methods and CAMIT for the word impromptu cybercrime euphemisms in the ICED dataset are shown in Table~\ref{tab:P_check}. 
Form the table, the existing methods for detecting word common cybercrime euphemisms have almost no detection capabilities for word improvised cybercrime euphemisms, especially Cant Reader\cite{Yuan2018ReadingTC}, X-phemisms\cite{Felt2020RecognizingEA}, EigenEuph\cite{Magu2018DeterminingCW},   GraphEuph\cite{Taylor2017SurfacingCH} and EPD\cite{Zhu2021EuphemisticPD} in the $top_{n}$ setting Under the experimental conditions of 5, 10, and 20, the obtained precision are almost 0. In comparison, MLM\cite{Zhu2021SelfSupervisedED} has improved slightly. However, our proposed method significantly outperforms the existing best methods. 
Especially under $top_{20}$ condition, compared with MLM\cite{Zhu2021EuphemisticPD}, the precision of our method is improved by 76 times, and the recall rate is increased by five times.

Figure~\ref{fig:boxplot} shows the prediction of impromptu euphemisms in ICED by two existing SOTA (MLM and EPD) and CAMIT. It is obvious that in CAMIT's prediction results, seed words(or phrases) rank significantly higher, indicating that CAMIT has a stronger ability to detect impromptu euphemisms.

\subsection{Ablation Study}
To quantitatively study the impact of context augmentation modeling (CAM), we conduct an ablation study, and results are shown in Figure \ref{fig:CAM}. After removing the CAM method, the precision of the model became lower, indicating that its ability to detect impromptu euphemisms weakened significantly. The recall increases continuously as the $\topp{k}$ increases, as is shown on the right side of Figure \ref{fig:CAM}. Also, without CAM, the model's precision rises up and drops down, indicating the model has reached its maximum ability, which proves that CAM improves the model greatly.

\begin{figure}[t]
  \centering
  \includegraphics[width=0.8\linewidth]{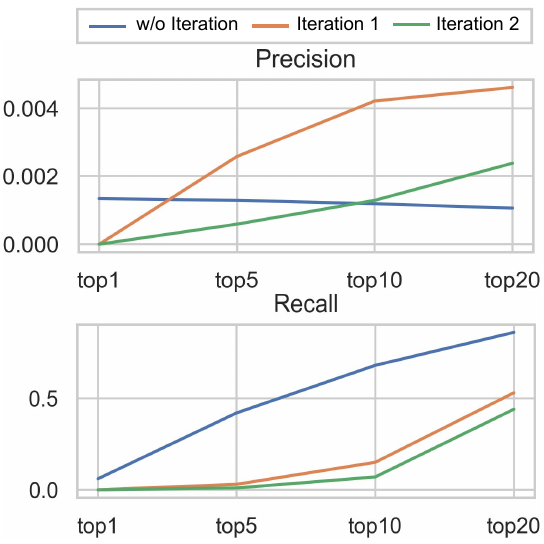}
  \caption{\small Precision and recall variation caused by different rounds of iterative training}
  \label{fig:ITER}
\end{figure}

Regarding the exploration of multi-round iteration training, as illustrated in Figure \ref{fig:ITER}, the experimental findings elucidate the nuanced impact of iteration rounds on model precision and recall. 
 We set up three sets of experiments, without iteration, iterating 1 round and iterating two rounds, in order to observe the changes in precision and recall.
The comparative analysis reveals that without multi-round iteration training, the model exhibits suboptimal performance, indicative of its limited capacity to accurately identify impromptu euphemisms. Furthermore, the diminishing returns observed with increasing iteration rounds suggest a threshold beyond which additional iterations yield diminishing improvements, signaling the onset of overfitting. Consequently, the findings advocate for a balanced approach without succumbing to the adverse effects of overfitting.

\subsection{Impact of Backbones}
\begin{table}[t]\small
\centering
\caption{\small Impact of different backbone of CAMIT. ``Prec'' and ``Rec'' denote ``Precion (\text{\textperthousand})'' and ``Recall'', respectively.}
\label{tab:backbone}
\setlength{\tabcolsep}{5pt}
\begin{tabular}{lcccccc}
\toprule
\multicolumn{1}{c}{\multirow{2}{*}{Backbone}} & \multicolumn{2}{c}{$\topp$5} & \multicolumn{2}{c}{$\topp$10} & \multicolumn{2}{c}{$\topp$20} \\ \cmidrule(lr){2-3} \cmidrule(lr){4-5} \cmidrule(lr){6-7} 
\multicolumn{1}{c}{}                        & Prec           & Rec          & Prec           & Rec           & Prec          & Rec          \\ \midrule
T5-small  & 0.80        & 0.04       & 0.95        & 0.58        & 0.70        & 0.62        \\
BERT-base      & 2.58        & 0.03       & 4.21        & 0.15        & 4.61        & 0.53        \\ \bottomrule
\end{tabular}
\end{table}
Our analysis in Table~\ref{tab:backbone} compared BERT and T5 backbones to determine their suitability for our methodology, focusing on precision and recall metrics. The findings indicate that BERT consistently outperforms T5 across all thresholds.

Key observations include:
1. BERT's higher precision and recall across the board signify its effectiveness in accurate prediction and comprehensive coverage.
2. The superior precision of BERT suggests its architecture, featuring bidirectional attention, is adept at grasping contextual nuances within our framework.
3. Although T5 shows comparable recall at lower thresholds, its precision lags behind BERT.

\subsection{Case Study}
Table~\ref{tab:goodsample} provides typical samples found during manual verification. It can be observed from the first two samples that some cybercrime euphemisms contain existing drug names or known euphemisms: ``mescaline'' in ``mescaline crystals''. Such samples can be classified using relatively simple word vector representations using Word2Vector. Moreover, sample 3 and 4 explanatory phrases for cybercrime euphemisms will also consist of words that do not contain known euphemisms at all, and their actual meaning can only be inferred by understanding contextual semantics. The last sample showed that there are some drug phrases that evolved from the chemical formula or molecular structure of drug molecules. 



\begin{table}[t]\small
\renewcommand\arraystretch{1.2}
\centering
\caption{\small Euphemisms confirmed during manual checks}
\label{tab:goodsample}
\begin{tabularx}{220pt}{X}
\hline \multicolumn{1}{c}{Samples} \\ \hline
\textbf{mescaline crystals} are probably the rarest available black market drug in existence you could do an extraction on san pedro cacti yourself    \\ \hline
... plays a heavy role since it is the most abundant  \textbf{cherry meth} of all the families and it play a huge role in the metabolism of many drugs ... \\ \hline
2g cocaine 5g colorado shatter 1 tab of ``double dipped'' no idea on the \textbf{exact potency} lucy = euphoric wonderland                               \\ \hline
rarest id say dmt mdma lsd and coke my \textbf{holy grail} is of course the rarest of what i can get dmtlsdmdma                                        \\ \hline
... \textbf{5-mapb} is a pretty good one very similar to mdma without the intense body load ...                                                        \\ \hline
\end{tabularx}
\end{table}

\section{Conclusion}
We mask a first exploration for impromptu cybercrime euphemism detection, collecting a dataset, and proposing appropriate evaluation metrics, which establish the foundation for this unexplored area. 
We propose a detection framework consisting of coarse- and fine-grained classification models. Moreover, we propose context augmentation and multi-round iterative training for the fine-grained classification model and conduct evaluation via ChatGPT. The experimental results show that the approach achieves a significant 76-fold improvement compared to previous SOTA methods.

\section*{Limitations}
Although our research has made progress, there are several limitations. First, to make a fair comparison with SOTA, we only trained language models with a comparable number of parameters. Second, our dataset was collected from a single online forum, which, while validating the effectiveness of our method, may potentially introduce bias. Lastly, since drug trafficking is considered a relatively common and serious form of cybercrime, this research focused on drug euphemisms only, which lacks diversity from a critical research perspective.

\section*{Acknowledgements}
This work was supported in part by the Major Key Project of PCL under Grant PCL2024AS103 and PCL2022A05, the Shenzhen Science and Technology Program under Grant RCJC20231211085918010, National Science Foundation of China (62476070), Shenzhen College Stability Support Plan (GXWD20231128103232001) and Department of Science and Technology of Guangdong (2024A1515011540).



\bibliography{custom}

\clearpage
\appendix
\onecolumn
\section{Full List of Euphemisms Generated by ChatGPT}
\label{sec:appendixA}

\begin{table}[h]
\centering
\caption{\small Euphemisms Generated by ChatGPT}
\begin{tabular}{r|p{11cm}p{11cm}p{11cm}p{11cm}}
\cline{1-2}
Drug names             & \multicolumn{1}{c}{Euphemisms}                                                                                                                                                                                                                                                                                               &  &  &  \\ \cline{1-2}
Alprazolam             & 1. Xanax 2. Bars 3. Zannies 4. Xanies 5. Blue footballs 6. Benzos 7. Upjohns 8. School buses 9. Handlebars 10. Planks 11. Bricks 12. Alpraz 13. Footballs 14. White boys 15. White girls 16. Bicycle parts 17. French fries 18. Yellow boys 19. Green monsters 20. Hulk bars                                                 &  &  &  \\ \cline{1-2}
Amphetamine            & 1. Speed 2. Uppers 3. Addys 4. Dexies 5. Bennies 6. Black beauties 7. Cross tops 8. Crank 9. Crystal 10. Glass 11. Ice 12. Meth 13. Go fast 14. Wake ups 15. Zoom 16. Amps 17. Pep pills 18. Study buddies 19. Smarties 20. Beans                                                                                            &  &  &  \\ \cline{1-2}
Cocaine                & 1. Coke 2. Blow 3. Snow 4. White 5. Powder 6. Yayo 7. Nose candy 8. Charlie 9. C 10. Dust 11. Toot 12. Line 13. Rail 14. Bump 15. Sniff 16. Skiing 17. Blizzard 18. Avalanche 19. Rockstar 20. Peruvian lady                                                                                                                 &  &  &  \\ \cline{1-2}
Ecstasy                & 1. E 2. X 3. Molly 4. Rolls 5. Beans 6. Love drug 7. Hug drug 8. Happy pill 9. Disco biscuit 10. Candy 11. Smarties 12. Skittles 13. Vitamin E 14. XTC 15. Adam 16. Eve 17. Scooby snacks 18. Thizz 19. Eskimo 20. White dove                                                                                                &  &  &  \\ \cline{1-2}
GHB                    & 1. G 2. Liquid E 3. Grievous Bodily Harm 4. Georgia Home Boy 5. Gina 6. Great Hormones at Bedtime 7. Goop 8. Gamma-OH 9. G-Juice 10. G-Riffick 11. Gook 12. G-Spot 13. GHB 14. Gammagoblin 15. G-Force 16. G-Bliss 17. G-Flower 18. G-Whiz 19. G-Thing 20. G-Phoria                                                          &  &  &  \\ \cline{1-2}
Heroin                 & 1. H 2. Smack 3. Dope 4. Junk 5. Horse 6. Black tar 7. China white 8. Brown sugar 9. Dragon 10. White lady 11. Big H 12. Skag 13. Thunder 14. Chiva 15. Boy 16. Girl 17. Mud 18. Poppy 19. Diesel 20. Snow                                                                                                                   &  &  &  \\ \cline{1-2}
Hydrocodone            & 1. Vikes 2. Norcos 3. Hydros 4. Watsons 5. Lorris 6. Tabs 7. 357s 8. Bananas 9. Fluff 10. Scratch 11. Juice 12. Hydro 13. Vicos 14. Whiteys 15. Watson 387 16. 10s 17. 5s 18. 7.5s 19. 10/325s 20. 5/325s                                                                                                                    &  &  &  \\ \cline{1-2}
Ketamine               & 1. K 2. Special K 3. Vitamin K 4. Kit Kat 5. Horse tranquilizer 6. Cat Valium 7. Super acid 8. K-hole 9. K-land 10. K-lean 11. K-powder 12. K-rave 13. K-rock 14. K-blast 15. K-bomb 16. K-juice 17. K-wax 18. K-dust 19. K-splash 20. K-loud                                                                                &  &  &  \\ \cline{1-2}
LSD                    & 1. Acid 2. Lucy 3. Tabs 4. Blotter 5. Doses 6. Trips 7. Hits 8. Paper 9. Alice 10. Sunshine 11. Electric Kool-Aid 12. Purple Haze 13. Orange Sunshine 14. White Lightning 15. Windowpane 16. Microdots 17. Sugar Cubes 18. Zen 19. Mind Candy 20. Heavenly Blue                                                              &  &  &  \\ \cline{1-2}
Marijuana Concentrates & 1. Wax 2. Shatter 3. Budder 4. Crumble 5. Honeycomb 6. Oil 7. Dabs 8. Rosin 9. Live resin 10. Sauce 11. Terp sauce 12. Diamonds 13. THCA crystals 14. BHO 15. CO2 oil 16. Distillate 17. Hash oil 18. Phoenix tears 19. Full melt 20. Bubble hash       
\\ \cline{1-2}
\end{tabular}
\end{table}

\begin{table}[t]
\centering
\caption{\small Euphemisms Generated by ChatGPT}
\begin{tabular}{r|p{11cm}p{11cm}p{11cm}p{11cm}}
\cline{1-2}
Drug names             & \multicolumn{1}{c}{Euphemisms}                                                                                                                                                                                                                                                                                               &  &  &  \\ \cline{1-2}
Mescaline              & 1. Mesc 2. Mescal 3. Mescalito 4. Peyote 5. Buttons 6. Cactus 7. San Pedro 8. Huachuma 9. Moon 10. God's flesh 11. Divine cactus 12. Sacred cactus 13. Visionary cactus 14. Shamanic medicine 15. Psychedelic cactus 16. Mind-expanding cactus 17. Spirit plant 18. Teacher plant 19. Wisdom plant 20. Hallucinogenic cactus &  &  &  \\ \cline{1-2}
Methamphetamine        & 1. Meth 2. Crystal 3. Ice 4. Tina 5. Crank 6. Glass 7. Speed 8. Go fast 9. Rocket fuel 10. Rocket candy 11. Rocket 12. Chalk 13. White cross 14. Zip 15. Zoom 16. Yaba 17. Batu 18. Shards 19. Hanyak 20. Hiropon                                                                                                            &  &  &  \\ \cline{1-2}
Methylphenidate        & 1. Ritalin 2. MPH 3. Kiddy Cocaine 4. Smarties 5. Vitamin R 6. Skippy 7. Diet Coke 8. West Coast 9. Pineapple 10. Kiddie Coke 11. R-ball 12. Rids 13. Ritz 14. Riddlin 15. Ritalina 16. Ritaline 17. Ritalyn 18. Ritalina LA 19. Ritalin SR 20. Ritalin LA                                                                   &  &  &  \\ \cline{1-2}
Opium                  & 1. O 2. Op 3. Poppy 4. Poppy seeds 5. Poppy straw 6. Poppy tea 7. Thebaine 8. Laudanum 9. Dreamer 10. Big O 11. Hop 12. Hophead 13. Auntie 14. Aunti Em 15. Aunti Emma 16. Black stuff 17. Block 18. Brown sugar 19. Chinese molasses 20. Dover's powder                                                                     &  &  &  \\ \cline{1-2}
Oxycodone              & 1. Oxy 2. OC 3. Hillbilly heroin 4. Percs 5. Blues 6. 30s 7. Roxies 8. Oxycontin 9. Oxy 80s 10. Oxy 40s 11. Oxy 20s 12. Oxy 10s 13. Oxy 5s 14. Oxy 15s 15. Oxy 30mg 16. Oxy 60mg 17. Oxy 120mg 18. Oxy 160mg 19. Oxy 240mg 20. Oxy 320mg                                                                                     &  &  &  \\ \cline{1-2}
PCP                    & 1. Angel dust 2. Rocket fuel 3. Hog 4. Sherm 5. Wack 6. Dust 7. Ozone 8. Embalming fluid 9. Supergrass 10. Killer weed 11. Love boat 12. Zoom 13. TAC 14. Rocket smoke 15. Crystal joint 16. Elephant tranquilizer 17. Happy sticks 18. Tic tac 19. Water 20. Peace pill                                                     &  &  &  \\ \cline{1-2}
Percocet               & 1. Percs 2. Paulas 3. Roxies 4. Blueberries 5. 512s 6. 30s 7. Oxy 8. Oxycontin 9. Oxycodone 10. Hillbilly heroin 11. Vikes 12. Painkillers 13. Happy pills 14. Killers 15. OCs 16. Oxy 80s 17. Oxy 40s 18. Oxy 20s 19. Oxy 10s 20. Oxy 5s                                                                                    &  &  &  \\ \cline{1-2}
Promethazine           & 1. Lean 2. Sizzurp 3. Purple drank 4. Dirty Sprite 5. Texas Tea 6. Barre 7. Purple jelly 8. Tsikuni 9. Drank 10. Syrup 11. Purple stuff 12. Leanin' 13. Purple rain 14. Purple oil 15. Lean syrup 16. Lean drink 17. Lean codeine 18. Lean promethazine 19. Lean cough syrup 20. Lean medication                             &  &  &  \\ \cline{1-2}
Psilocybin Mushrooms   & 1. Shrooms 2. Magic mushrooms 3. Caps 4. Boomers 5. Blue meanies 6. Liberty caps 7. Gold caps 8. Philosopher's stones 9. Mushies 10. Funguys 11. Zoomers 12. God's flesh 13. Sacred mushrooms 14. Teonanácatl 15. Psilocybe 16. Psilocin 17. Psilo 18. Little smoke 19. Silly putty 20. Alice in Wonderland                  &  &  &  \\ \cline{1-2}
Steroids               & 1. Roids 2. Juice 3. Gear 4. Sauce 5. Pumpers 6. Stackers 7. Hype 8. A-bombs 9. D-bol 10. Winnie 11. Tren 12. Deca 13. Test 14. Var 15. Clen 16. GH 17. ECA 18. Nolva 19. Arimidex 20. Proviron                                                                                                                              &  &  &  \\ \cline{1-2}
Synthetic Cathinones   & 1. Bath salts 2. Flakka 3. Ivory Wave 4. Cloud Nine 5. Vanilla Sky 6. White Lightning 7. Scarface 8. Hurricane Charlie 9. Lunar Wave 10. Bliss 11. Blue Silk 12. Purple Wave 13. Red Dove 14. Snow Leopard 15. Stardust 16. White Dove 17. White Rush 18. White Sands 19. Zoom 20. Charge                                    &  &  &  \\ \cline{1-2}
\end{tabular}
\end{table}

\end{document}